\documentclass[final,5p,times,twocolumn,authoryear]{elsarticle}

\usepackage{amssymb}
\usepackage{amsmath}
\usepackage[utf8]{inputenc}

\usepackage{caption}
\usepackage{float} 
\usepackage[colorlinks=true, allcolors=blue]{hyperref}

\begin{document}

\begin{frontmatter}
\author[a]{Shahbaz Alvi\corref{cor1}}
\ead{shahbaz.alvi@cmcc.it}
\author[a]{Giusy Fedele}
\author[a,c]{Gabriele Accarino}
\author[a,b]{Italo Epicoco}
\author[a]{Ilenia Manco}
\author[a]{Pasquale Schiano}
\cortext[cor1]{Corresponding author}

\affiliation[a]{organization={CMCC Foundation - Euro-Mediterranean Center on Climate Change},
            city={Lecce},
            postcode={73100},
            country={Italy}}

\affiliation[b]{organization={University of Salento,Department of Innovation Engineering},
            city={Lecce},
            postcode={73100},
            country={Italy}
            }
\affiliation[c]{organization={Columbia University, Department of Earth and Environmental Engineering},
            city={New York},
            country={USA}
            }
\title{\texttt{OpFML}: Pipeline for ML-based Operational Inference} 

\begin{abstract}
Machine learning models for climate and Earth science are becoming increasingly capable, yet model deployment into operational use remains a largely unaddressed challenge: general-purpose model-serving tools, such as MLflow and KServe, assume input data availability at the inference node, while data acquisition, failure handling, and preprocessing are trusted to a separate workflow. We present \texttt{OpFML} (\textbf{Op}erational \textbf{F}orecasting with \textbf{M}achine \textbf{L}earning) - a configurable pipeline integrating the four steps of operational inference into a single TOML-configured workflow: data consumption, contingency handling, preprocessing, and model inference. By consolidating these steps, \texttt{OpFML} removes the significant boilerplate code required for each new deployment. We demonstrate the pipeline on the operational forecasting of daily fire activity over southern Italy.
\end{abstract}


\end{frontmatter}



\section{Introduction}\label{sec:intro}
Machine learning (ML) methods support research and operations in the field of Climate and Earth Science by leveraging increasingly precise measurements from both in-situ instruments and remote sensing (\cite{rev_ml_1, rev_ml_2}). Considerable efforts have been devoted to machine learning approaches for various use cases in the field of climate; however, such applications remain largely at the proof-of-concept stage. While a plethora of tools are available to support training and testing of ML models (e.g. \cite{Paszke2019_PyTorch,tensorflow2016_whitepaper,kit4dl}), there is a lack of tools to fulfill the specialized needs of the climate community in operational inference where data-pipeline orchestration is equally important as model inference execution: autonomous ingestion of heterogeneous covariates, contingency handling when upstream sources fail, configurable per-variable preprocessing, and model inference. General-purpose ML serving frameworks (e.g., BentoML \cite{BentoML}, KServe \cite{KServe}, MLflow \cite{Zaharia2018_MLflow}, TorchServe \cite{TorchServe}) assume data readiness at the inference node, delegating the aforementioned steps to separate workflows, resulting in a potential gap in operational ML applications across several use cases in Earth-science domains. 

In the explicit use case of wildfire forecasting \cite{alonso2003} presented an operational system for fire risk associated with a given day over Galicia in north-west Spain. More recently, (\cite{DiGiuseppe2025}) presented results from ECMWF’s Probability of Fire (PoF) system for daily global wildfire forecasts. However, a freely available, configurable pipeline code is currently lacking.

This manuscript presents \texttt{OpFML}, an operational, modular, adaptable, and flexible pipeline for downstream operational inference using a PyTorch Lightning ML model for the climate community \footnote{
https://github.com/CMCC-Foundation/opfml}. OpFML is an integrated TOML-configured workflow that unifies the four essential steps for operational forecasting. The integrated ML model is decoupled from the pipeline, enabling different models to be served across geographic regions with minimal configuration changes. The containerised pipeline can be deployed on orchestration platforms such as Kubernetes, for autonomous periodic execution to support operational management.

\section{Pipeline Layout}\label{sec:pip_layout}
\subsection{Structure of the pipeline}\label{subsec:pip_strut}
The pipeline is organised around a small set of classes, each responsible for one step in the operational inference workflow. Their roles are summarised in Table~\ref{tab:classes} and described in detail in the following subsections.

\begin{table}
\centering
\caption{Principal classes of the OpFML pipeline and their responsibilities.}
\label{tab:classes}
\begin{tabular}{@{}p{0.34\linewidth}p{0.58\linewidth}@{}}
\textbf{Class} & \textbf{Responsibility} \\
\texttt{DataStore} &
Consumes covariates from the upstream sources configured. \texttt{CriticalFailure\-Contingency}, in the \texttt{DataStore} class, applies contingency policy for a covariate when upstream data are unavailable. \\[2pt]
\texttt{PreProcessing} &
Applies the per-variable transformations at runtime, in FIFO order, before inference. \\[2pt]
\texttt{InferenceModule} &
Performs inference on the preprocessed data and stores the results. \\
\end{tabular}
\end{table}
\subsubsection{Layout}\label{subsubsec:pip_layout}
\begin{figure}[ht]
    \centering
    \includegraphics[width=\linewidth]{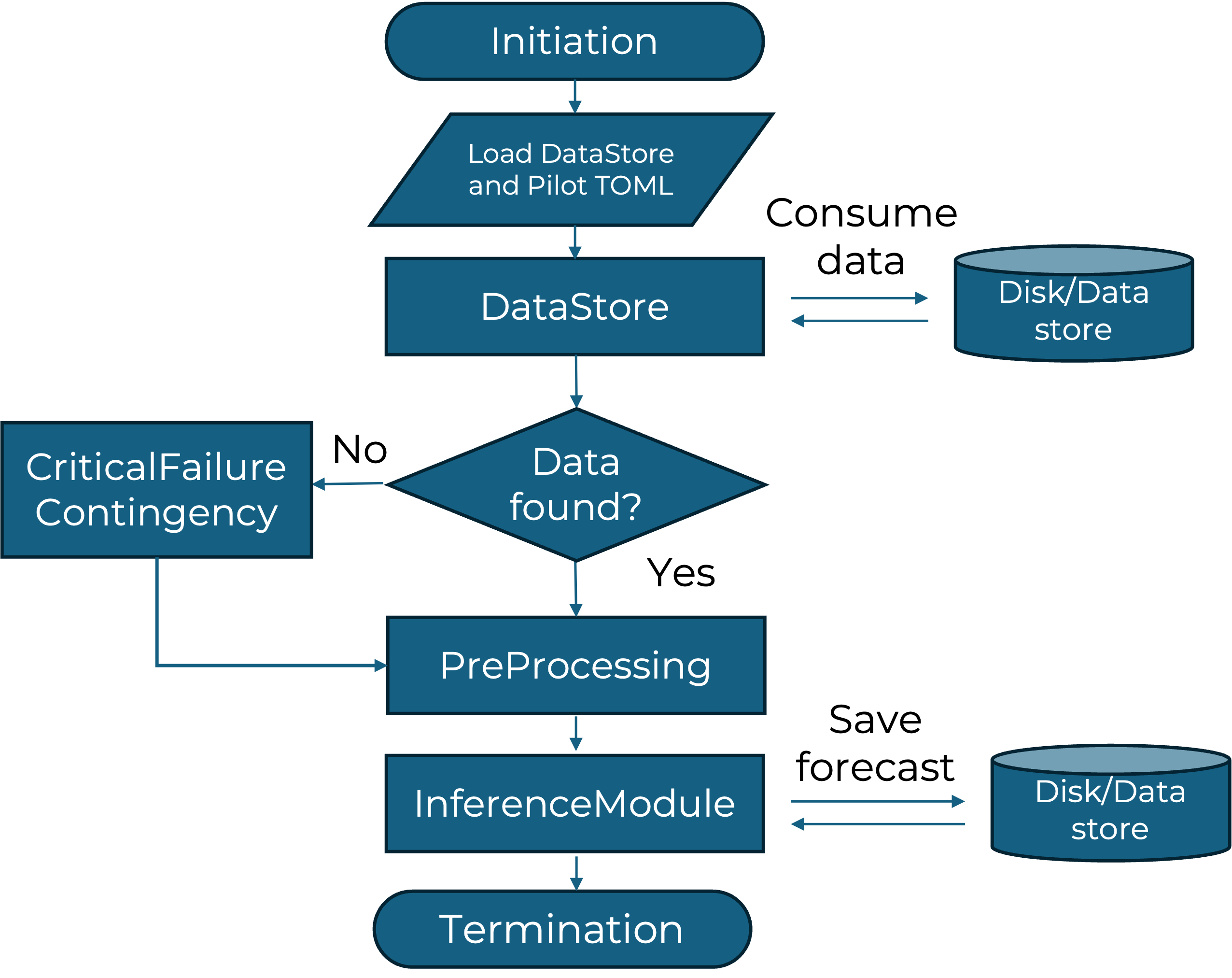}
    \caption{Schematic flow chart of the pipeline. }
    \label{fig:fdi_pipeline_flow}
\end{figure}

The pipeline has three main components: \texttt{DataStore}, \texttt{PreProcessing}, and \texttt{InferenceModule} for data consumption, preprocessing, and model inference, respectively. Contingency handling is a fourth functional step, implemented within DataStore. Two TOML (Tom’s Obvious Minimum Language) configuration files are required to execute the pipeline: one for the data store configuration and one for the operational inference over a geographic area (called the pilot TOML file). The human-readable key–value format of TOML files provides a high level of configurability, allowing users to modify and adapt the pipeline without modifying the source code. The pipeline layout is summarily depicted in Figure \ref{fig:fdi_pipeline_flow}.

\subsubsection{Data consumption and data store TOML file}\label{subsubsec:pip_ds_class}

\begin{figure}
    \centering
    \includegraphics[width=1\linewidth]{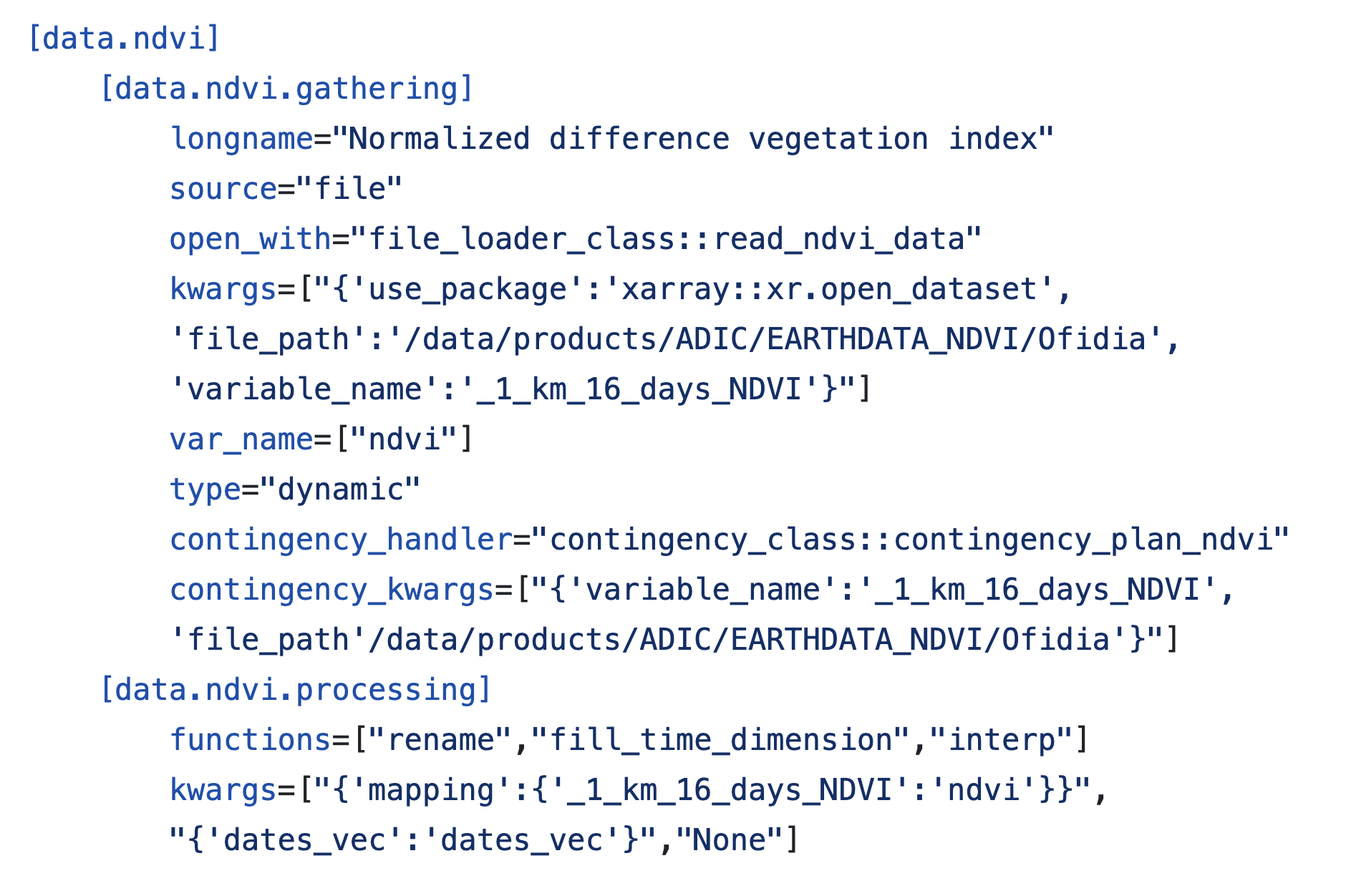}
    \caption{Configuration for consuming the NDVI variable in the data store TOML file. The \texttt{DataStore} class utilizes the configuration in the ``gathering'' section to consume the data. The transformations to be applied are described in the ``processing'' section together with the input required by each function.}
    \label{fig:toml_dds}
\end{figure}

Downstream inference tasks in climate and Earth science require consumption of the most recent, task-specific covariates. The \texttt{DataStore} class is responsible for consuming the covariates according to the configured methods (could be a file or an API, e.g., NASA Appears or CMCC DDS). Each covariate is configured as a single entry in the data store TOML file,
organised into two blocks that reflect the two purposes of the file:
\begin{itemize}
    \item \textbf{Gathering}: How the covariate is acquired. This includes the loader function (\texttt{open\_with}) and its arguments (\texttt{kwargs}), the variable name, and the static/dynamic flag. For dynamic covariates, it also declares the contingency handler (\texttt{contingency\_handler}) and its arguments (\texttt{contingency\_handler\_kwargs}), invoked when the latest data cannot be consumed from the upstream source (and the function declared in \texttt{open\_with} returns \texttt{None}).
    \item \textbf{Processing}: The ordered list of transformations (\texttt{functions}) applied to the covariate before model inference (see Section~\ref{subsubsec:preprocess}), each with its own arguments.
\end{itemize}
Figure~\ref{fig:toml_dds} shows the configuration of the NDVI covariate as an example. A sample data store file is provided in the supplementary material.

The unavailability of upstream data is a significant bottleneck for autonomous downstream applications, as the critical failure cascades to downstream workflows unless these scenarios are manually mitigated to synchronize the workflows. The \texttt{CriticalFailureContingency}, in the \texttt{DataStore} class, implements the contingency plan in the case the data consumption fails, allowing the user to integrate the mitigation strategy to avoid critical failure of the pipeline, as depicted in Figure \ref{fig:fdi_pipeline_flow}. In the excerpt shown in Figure \ref{fig:toml_dds}, the primary function for consuming the latest NDVI data is \texttt{read\_ndvi\_data}. In the event of failure, \texttt{read\_ndvi\_data} returns \texttt{None}, the function declared in the parameter \texttt{contingency\_handler} is called using parameters in \texttt{contingency\_handler\_kwargs}.

\subsubsection{Data pre-processing}\label{subsubsec:preprocess}
\begin{figure}
    \centering
    \includegraphics[width=1\linewidth]{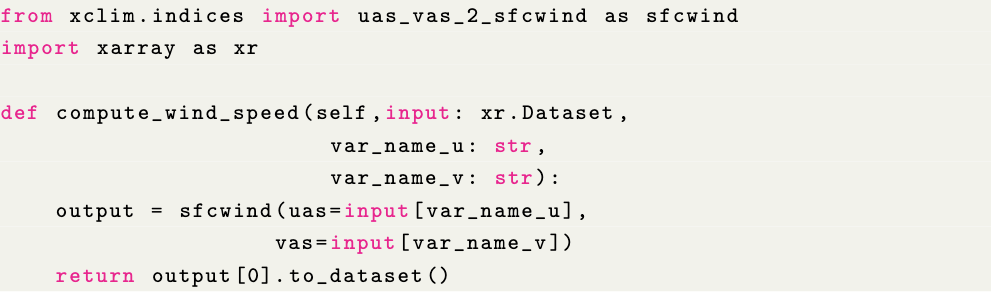}
    \caption{Code snippet of the transformation function in \texttt{PreProcessingUtils} class for computing the wind speed using library \texttt{xclim}, and function \texttt{uas\_vas\_2sfcwind}. The input \texttt{xarray} Dataset is expected to contain the wind speed components.}
    \label{code:windspeed}
\end{figure}

Some consumed data may require preprocessing before passing it to the ML model for inference. In the provided code, the \texttt{PreProcessingUtils} class, declared in the pilot TOML file, serves as the library of transformations that are sequentially applied by the \texttt{PreProcessing} class at runtime (first transformation is applied first). An example \texttt{Python} function for calculating wind speed from the consumed \texttt{u} and \texttt{v} velocity components is illustrated in Figure \ref{code:windspeed}. Schematically, the \texttt{DataStore} class consumes the variable, either directly or via the contingency plan for this variable, passes the consumed variables to this function, which outputs the wind speed, and returns it as the output. 

\subsubsection{Pilot TOML file and model inference}
The pilot TOML file stores the configuration for a specific pilot site where model inference needs to be performed. The parameters are grouped into three categories:
\begin{itemize}
    \item \textbf{Model parameters}: the path to the saved ML model checkpoint and the inputs required to initialise it for inference as required by the inference module.
    \item \textbf{Pipeline parameters}: the \texttt{DataStore} class to be used for this pilot and the covariates to be consumed among those configured in the data store TOML.
    \item \textbf{Spatial parameters}: the geographic bounding box of the pilot site.
\end{itemize}
Figure \ref{fig:pilot_info} shows an excerpt from a pilot TOML file; the complete file is provided in the supplementary material. All consumed covariates, after being preprocessed according to their respective configurations, are passed to the inference module, which performs inference on the data, and the inference result is stored. 

\begin{figure}[ht]
    \centering
    \includegraphics[width=1\linewidth]{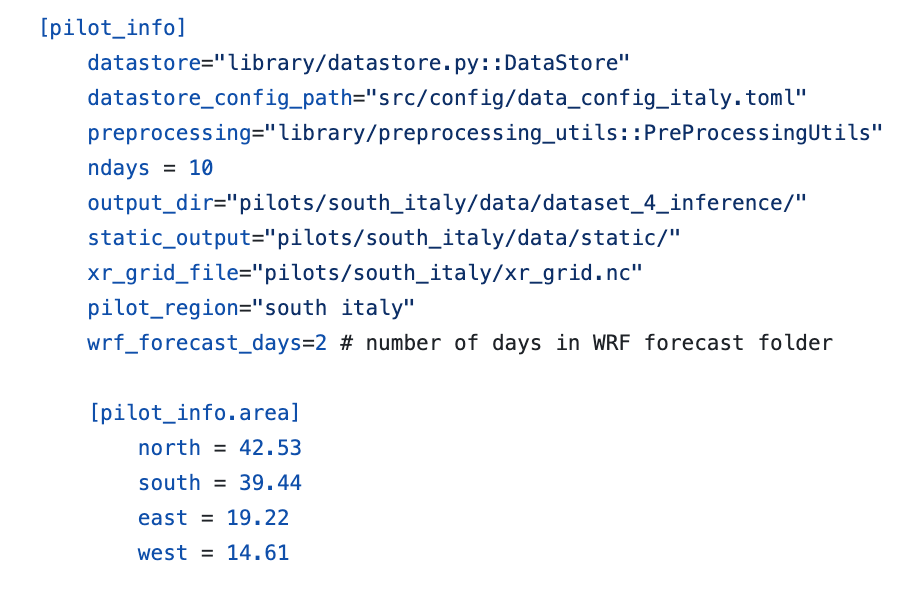}
    \caption{ A section of the pilot TOML file describing the data store class and processing class to be used for the pilot site. Additionally, parameters can also be defined depending on the use case.}
    \label{fig:pilot_info}
\end{figure}

\subsection{Adaptability of the pipeline}\label{subsec:pip_robust}
The pipeline was designed keeping high-level configurability and adaptability in mind to serve use cases beyond the one demonstrated in this work. Several configurable elements of the workflow are described here. \textbf{Model substitution.} Integrating a different PyTorch Lightning ML model requires only updating the path to the PyTorch checkpoint in the pilot TOML file. Different pilot TOMLs can specify different models, enabling pilot-specific models fine-tuned for the particular geographical region. \textbf{Data store substitution.} The \texttt{DataStore} class for a given pilot is declared in the pilot TOML file (Figure \ref{fig:pilot_info}). Pilot-specific upstream sources can therefore be integrated in customized \texttt{DataStore} classes and declared in the pilot TOML. \textbf{Preprocessing library substitution.} The transformation library used by \texttt{PreProcessing} is also configurable through the pilot TOML file. This allows pilots to maintain their own libraries of transformation functions, tailored to the pilot-specific predictors and conventions. Moreover, data transformations are straightforward to configure through a well-recognized Python dictionary format. \textbf{Contingency handling.} The proposed pipeline integrates both data consumption and mitigation strategy for critical failure into a single workflow, avoiding the manual overhead required to maintain and synchronize independent workflows.  

\section{The daily fire activity use case}\label{sec:use_case}
We demonstrate the pipeline's functionality for operational forecasting of daily fire activity in southern Italy. The ML architecture, training dataset, and training strategy are as described in \citet{Kondylatos2022}; the model outputs the
probability of each pixel belonging to the \textit{Fire} or \textit{No-fire} category. The two TOML files (Section~\ref{subsubsec:pip_layout}) are configured for this use case as follows: the data store TOML declares the 24 covariates listed in Table~\ref{tab:fire_preds_app}, acquired through the CMCC Foundation's Data Delivery
Service (\cite{CMCC_DDS}); the pilot TOML specifies the bounding box covering southern Italy (Figure~\ref{fig:pilot_info}) and other information required for inference. Both files are provided in the supplementary material to reproduce the maps in Section~\ref{sec:results}.

\subsection{Forecast model input}
In the following subsections, we describe the dataset used for the daily fire activity forecast.\subsubsection{Numerical models for weather variables}\label{subsubsec:forecast}
Short-range high-resolution weather forecasts used for wildfire risk assessment were generated using version 4.2.1 of the Weather Research and Forecasting (WRF) model (\cite{Skamarock2021_WRFv4.3}) at $\approx$2 km horizontal resolution. Configuration details and parameter choices are reported in \cite{Manco2023_ComparativeWRF}. NCEP’s Global Forecast System (GFS) was selected for operational implementation following a comparative assessment of GFS and Integrated Forecasting System (IFS) forcing. Gridded and in-situ observational sources were used to perform validation over the pilot sites on a case-by-case basis. Further details of the operational implementation and validation are provided in the supplementary documents. 
\subsubsection{Non-weather variables}
The non-weather variables used to forecast fire activity, the parent repository from which they were acquired, and the pre-processing applied to these variables are listed in Table \ref{tab:fire_preds_app}.

\subsection{Running the pipeline}
The following command initiates the workflow for computing the fire activity map for 2024-07-09. 
\begin{verbatim}
    python src/init.py
    --conf=pilots/south_italy/setup.toml
    --date 2024-07-09 --collect_data 
    --prepare_static --make_forecast --netcdf  
\end{verbatim}
The workflow differentiates between static and dynamic variables, since static variables can be consumed and stored once locally and used repeatedly. \texttt{prepare\_static} instructs the pipeline to consume static variables, while \texttt{collect\_data} option instructs the workflow to consume the dynamic dataset. \texttt{netcdf} option instructs the pipeline to save model output both as \texttt{netCDF} and \texttt{GeoJSON} format.

The model input data used to generate the maps presented in Section \ref{sec:results} are available in the repository. The following command runs the inference module with the stored dataset 

\begin{verbatim}
    python src/init.py
    --conf=pilots/south_italy/setup.toml
    --date 2024-07-09 --make_forecast  --netcdf
\end{verbatim}

\subsection{Forecast maps from the pipeline}\label{sec:results}
\begin{figure}[ht]
    \centering
    \begin{minipage}[b]{0.48\textwidth}
        \centering
        \includegraphics[width=\textwidth]{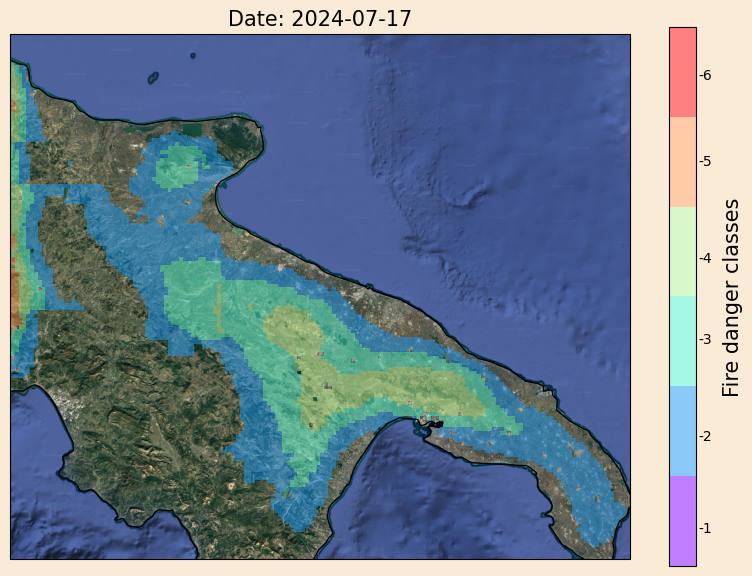}
    \end{minipage}
    \begin{minipage}[b]{0.48\textwidth}
        \centering
        \includegraphics[width=\textwidth]{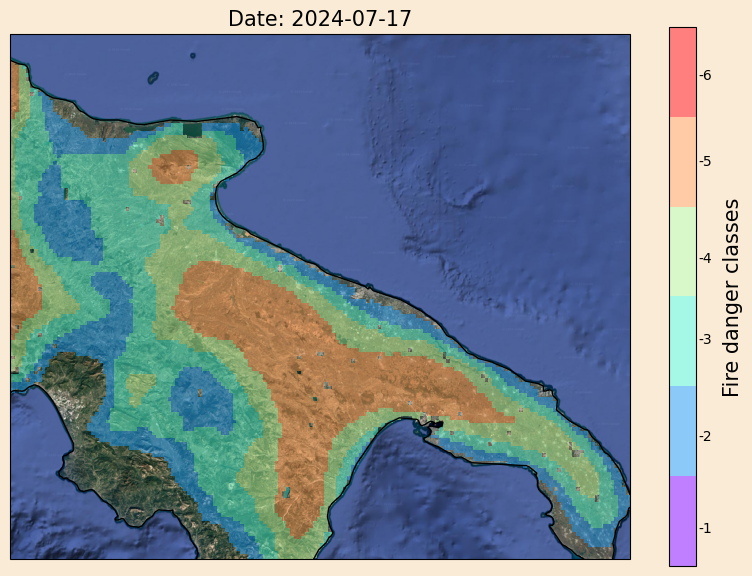}
    \end{minipage}
    \caption{Daily fire activity forecast over southern Italy for some days in July of 2024. A fire icon marks the confirmed fire events on that day.}
    \label{fig:italy_fdi}
\end{figure}

The wildfire activity maps produced by the pipeline are shown here for some days in July 2024 for demonstration purposes in Figure \ref{fig:italy_fdi}. The operational forecast is reported in 6 danger classes - Category 1 is the lowest fire danger category, which is suppressed for better visualization, and Category 6 is the highest probability of fire occurrence. 
The maps produced downstream of the pipeline serve as the operational fire-activity forecast and, in turn, provide the basis for the separate validation of the underlying ML model, which is beyond the scope of this work.

\section{Discussion and conclusion}
Producing a periodic, operational inference for a target variable with a trained ML model is rarely a single step. It requires consuming the most recent covariates from upstream sources, handling missing-data contingency, preprocessing each variable, and finally executing inference. Assembling such a workflow from scratch is laborious, and few off-the-shelf solutions address it as an integrated whole.

OpFML addresses this gap by integrating these four steps into a single, configurable workflow. Critically, the pipeline is modular: each step is configured independently through the TOML files, so a model, a data source, a preprocessing routine, or a contingency policy may be substituted without altering the others or the source code. By consolidating this orchestration, OpFML removes the substantial
boilerplate code each new deployment would otherwise require, allowing a trained model to be brought into operational use through configuration changes rather than new code.

The pipeline is implemented entirely in \texttt{Python}, the de facto language for ML applications, and serves models trained with \texttt{PyTorch Lightning}. We demonstrated its functionality on the operational forecasting of daily fire activity over southern Italy. The pipeline was developed for the SILVANUS project \citep{SILVANUS2024}, a Horizon 2020 Green Deal project, to produce fire activity
forecasts for the pilot sites.

The current implementation has two practical limitations. First, the served model must be built with \texttt{PyTorch Lightning}; models trained in other frameworks are not yet supported. Second, the pipeline primarily supports datasets in the netCDF format; although netCDF is standard in the climate and Earth-science community, upstream sources often distribute data in other formats, which requires adapting the \texttt{DataStore} class accordingly. Both constraints reflect deliberate scoping rather than architectural limits.

Although demonstrated here on the daily fire-activity forecast, the modular design of OpFML provides a clear path to broaden framework and format support, and to generalise to other operational prediction tasks in climate and Earth science.

\section{Acknowledgments}
The authors acknowledge, with gratitude, the constant support of the development team of the CMCC DDS service (Gabriele Tramonte and Valentina Scardigno) throughout the development of this work. This project is partially funded by the European Union’s Horizon 2020 research and innovation program under grant agreement No 101037247.

\bibliographystyle{elsarticle-num-names} 
\bibliography{references}

\newpage
\appendix
\section{Supplementary information}
\subsection{List of fire predictors}
The Table \ref{tab:fire_preds_app} shows the list of fire predictors used in the use cases, along with the variables, and their parent repository is shown as well. 
\begin{table}[ht]
    \centering
    \begin{tabular}{|c|c|c|c|}
        \hline
        \textbf{No.} & \textbf{Variable} & \textbf{Source} & \textbf{Aggregation} \\
        \hline
         1 & \begin{tabular}{@{}c@{}}Normalized  \\ difference \\ vegetation  \\ index \\ (NDVI) \end{tabular} & \begin{tabular}{@{}c@{}}16-daily, \\ NASA MODIS \\ Terra \cite{ds_ndvi_Didan2021_MODIS_VI} \end{tabular}  & \begin{tabular}{@{}c@{}}16-daily, \\ forward filled \\ to  1-day \\ resolution.  \end{tabular} \\
        \hline
         2-3 & \begin{tabular}{@{}c@{}}Daily LST \\ day \& night  \end{tabular} & \begin{tabular}{@{}c@{}}NASA MODIS \\  \cite{Wan2015_MOD11} \end{tabular} & - \\
        \hline
        4 & \begin{tabular}{@{}c@{}}Dew point  \\  temperature \\ at 2-meter \end{tabular} & \begin{tabular}{@{}c@{}}Details in \\ Section \ref{subsubsec:forecast} \end{tabular}   & \begin{tabular}{@{}c@{}}Hourly \\ to \\  1-day max \end{tabular} \\
        \hline
         5 & \begin{tabular}{@{}c@{}}Air  \\  temperature \\ at 2-meter \end{tabular} & \begin{tabular}{@{}c@{}}Details in \\ Section \ref{subsubsec:forecast} \end{tabular} & \begin{tabular}{@{}c@{}}Hourly \\ to \\  1-day max \end{tabular} \\
        \hline
        6 & \begin{tabular}{@{}c@{}}Surface  \\  pressure \end{tabular}   & \begin{tabular}{@{}c@{}}Details in \\ Section \ref{subsubsec:forecast} \end{tabular} & \begin{tabular}{@{}c@{}}Hourly \\ to \\  1-day max \end{tabular} \\
        \hline
        7  & \begin{tabular}{@{}c@{}}Total  \\  precipitation \end{tabular}   & \begin{tabular}{@{}c@{}}Details in \\ Section \ref{subsubsec:forecast} \end{tabular} & \begin{tabular}{@{}c@{}}Hourly \\ to \\  1-day max \end{tabular}  \\
        \hline
         8 & \begin{tabular}{@{}c@{}}u\&v wind \\ speed \\ components \\ (10m)  \end{tabular} & \begin{tabular}{@{}c@{}}Details in \\ Section \ref{subsubsec:forecast} \end{tabular} & \begin{tabular}{@{}c@{}}Hourly \\ to \\  1-day max \end{tabular} \\
        \hline
         9 & \begin{tabular}{@{}c@{}}Relative \\ humidity \end{tabular}   & \begin{tabular}{@{}c@{}}Details in \\ Section \ref{subsubsec:forecast} \end{tabular} & \begin{tabular}{@{}c@{}}Hourly \\ to \\  1-day min \end{tabular} \\
        \hline
         10 & \begin{tabular}{@{}c@{}}Elevation \\ (DEM) \end{tabular} & \begin{tabular}{@{}c@{}} Copernicus \\ \cite{CopernicusDEM2023} \end{tabular}  & - \\
        \hline
         11 & Slope & \begin{tabular}{@{}c@{}}Derived  \\ from DEM  \end{tabular}  & - \\
        \hline
         12 & \begin{tabular}{@{}c@{}}Distance to  \\ roads \end{tabular} & \begin{tabular}{@{}c@{}} WorldPop.org \\ \cite{Lloyd2019_GlobalPop} \end{tabular} & - \\
        \hline
         13 & \begin{tabular}{@{}c@{}}Distance to  \\ waterway \end{tabular}  & \begin{tabular}{@{}c@{}} WorldPop.org \\  \cite{Lloyd2019_GlobalPop} \end{tabular} & - \\
        \hline
         14 & \begin{tabular}{@{}c@{}}Population  \\ density \end{tabular}   & \begin{tabular}{@{}c@{}} WorldPop.org \\ \cite{Lloyd2019_GlobalPop} \end{tabular}   & - \\
        \hline
         15-24 & \begin{tabular}{@{}c@{}}Remapped \\ Corine Land \\ Cover \end{tabular} &  \begin{tabular}{@{}c@{}} Copernicus \\ \cite{Copernicus_CLC_2018} \end{tabular} & - \\
        \hline
    \end{tabular}
    \caption{List of fire predictors used in the use cases, their source, and the summary statistics applied to them.}
    \label{tab:fire_preds_app}
\end{table}
\subsection{WRF weather forecast}
\subsubsection{Operational workflow for WRF weather forecast}
Short-range high-resolution weather forecasts used for wildfire risk assessment were generated using version 4.2.1 of the Weather Research and Forecasting (WRF) model (\cite{Skamarock2021_WRFv4.3}). 
More specifically, within this study, we employ a version of the WRF model configured at $\approx$2 km horizontal resolution ((\cite{Manco2023_ComparativeWRF}), hence referred to as WRF\_2km@CMCC). Configuration details and the sensitivity analysis that guided parameter choices are reported in WRF\_2km@CMCC (also summarized in Table \ref{tab:wrf_config}).
The forecasting system is configured to produce daily operational 72-hour forecasts at hourly resolution, providing the set of variables required for fire-danger computations (see Appendix, Table \ref{tab:fire_preds_app}) over southern Italy. Two global forcings were tested for the limited-area WRF setup: NCEP’s Global Forecast System (GFS) and ECMWF’s Integrated Forecasting System (IFS). An in-depth evaluation was first performed with IFS-forced runs over the Apulia region (April–October, 2019–2020), followed by a comparative assessment of GFS- and IFS-forced outputs. A preliminary comparison showed broadly comparable performance between the two forcings; because IFS boundary data are routinely available with an approximate three-day delay, GFS was selected for operational implementation.
\begin{table}[ht]
    \centering
    \begin{tabular}{|c|c|}
    \hline
        \textbf{Model} & \textbf{WRF (ARW)} \\
    \hline
        Version & WRF v4.2.1 \\ 
    \hline
        Boundary forcing & \begin{tabular}{@{}c@{}}IFS ECMWF 0.05° ($\simeq 6$ km) \\ NCEP GFS 0.25° ($\simeq 27.8$ km) \end{tabular} \\
    \hline
        \begin{tabular}{@{}c@{}}Lateral Boundary  \\ Condition (LBC) \end{tabular}   & Update frequency 1 h \\
    \hline
        Soil initializations & \begin{tabular}{@{}c@{}}Temperature and moisture   \\ were obtained by \\ interpolation from GFS \end{tabular} \\
    \hline
        Horizontal resolution & \begin{tabular}{@{}c@{}}0.018° (2 km), 0.01°($\simeq 1.7$ km) \end{tabular}\\
    \hline
        Time step & 12 seconds\\
    \hline
        N° vertical levels & 60\\
    \hline
        Output frequency & 1 hour\\
    \hline
        Coordinate system & \begin{tabular}{@{}c@{}}Horizontal: Conformal   \\ Lambert on a \\ latitude‑longitude  \\ grid vertical: \\ sigma-pressure  \end{tabular} \\
    \hline
    Radiation scheme & RRTMG \cite{Baek2017_RevisedRadiation} \\
    \hline
    \begin{tabular}{@{}c@{}}Non-orographic gravity  \\ wave drag \end{tabular} & \begin{tabular}{@{}c@{}}The scheme includes  \\  two subgrid topography \\ effects: \\ gravity wave drag and \\ low-level flow blocking \\ \cite{Lott1997_OrographicDrag}  \end{tabular} \\
    \hline
    \begin{tabular}{@{}c@{}}Sub-grid scale \\ orographic drag \end{tabular} & \begin{tabular}{@{}c@{}}SSO scheme \\ \cite{Lott1997_OrographicDrag} \end{tabular}   \\
    \hline
    Microphysics & Morrison 2-moment \\
    \hline
    Convection scheme & \begin{tabular}{@{}c@{}}No cumulus \\ parameterization \end{tabular}   \\
    \hline
    Turbulent transfer & \begin{tabular}{@{}c@{}}Yamada-Joshi scheme \\ \cite{Yamada1989_SecondOrder} \end{tabular} \\
    \hline
    Land surface scheme & \begin{tabular}{@{}c@{}}Simplified Simple Biosphere   \\ Model \cite{Xue1991_SSiB}  \end{tabular}   \\
    \hline
    Land use dataset & \begin{tabular}{@{}c@{}}GlobCover2009 \\ \cite{Arino2012} \end{tabular}    \\
    \hline
    Time integration & 3rd‑order Runge Kutta \\
    \hline
    Horizontal grid & Arakawa C staggering \\
    \hline
    Spatial discretization & 6th‑order centered differences\\
    \hline
    \end{tabular}
    
    \caption{Key features of WRF\_2km@CMCC configuration}
    \label{tab:wrf_config}
\end{table}

\subsubsection{Validation}
Key diagnostics, hourly 2-m temperatures (max/mean/min), 10-m wind speed, and total daily precipitation, were compared against available reference data depending on the study area. Validation combined standard statistical diagnostics (e.g., mean bias, rmse; here not shown) with targeted temporal and spatial comparisons to capture both variability and patterns’ distribution. Across the test sites, the GFS-driven simulations demonstrated good performance compared to observational data and adequate performance for fire-weather and fire-risk applications, supporting their use in the operational workflow.
Over Italy, the following datasets have been used: (i) the E-OBS (\cite{Cornes2018_EOBS}) gridded observational dataset providing daily atmospheric variables at $\approx$ 11 km resolution, based on in situ measurements; (ii) in situ weather stations provided by institutions (Agenzia Regionale Attività Irrigue e Forestali - ARIF); (iii) the dataset VHR-REA\_IT, a high-resolution dataset derived from dynamically downscaling ERA5 with the COSMO-CLM model, offering hourly data at 2.2 km resolution over the Italian Peninsula (\cite{Raffa2021_VHR_REA_IT,Raffa2023_VHRProjections}). Model performance was statistically evaluated, demonstrating comparable performance between GFS- and IFS-driven simulations. and strong agreement between WRF\_2km@CMCC and the reference datasets (not shown here), supporting its use in fire danger and fire weather risk assessments. The results related to the model skills for both configurations (IFS- and GFS-driven) over the past period demonstrated that both configurations may be used for operational purposes.
\end{document}